\documentclass[10pt,twocolumn,letterpaper]{article}

\usepackage{3dv}
\usepackage{times}
\usepackage{epsfig}
\usepackage{graphicx}
\usepackage{amsmath}
\usepackage{amssymb}
\usepackage{float}
\usepackage[ruled,linesnumbered,commentsnumbered]{algorithm2e}
\SetKwInput{KwInput}{Input}                
\SetKwInput{KwOutput}{Output}
\SetKw{KwIn}{in}
\SetKw{KwOr}{or}
\usepackage{array}
\usepackage{booktabs}
\usepackage{amssymb}% http://ctan.org/pkg/amssymb
\usepackage{pifont}% http://ctan.org/pkg/pifont
\newcommand{\xmark}{\ding{53}}

\threedvfinalcopy % *** Uncomment this line for the final submission

\def\threedvPaperID{32} % *** Enter the 3DV Paper ID here

\ifthreedvfinal\fi
\begin{document}

%%%%%%%%% TITLE
\title{Open-set 3D Object Detection}

\author{Jun Cen \quad Peng Yun \quad Junhao Cai \quad Michael Yu Wang \quad Ming Liu\\
The Hong Kong University of Science and Technology\\
{\tt\small \{jcenaa, pyun, jcaiaq\}@connect.ust.hk, \{mywang, eelium\}@ust.hk}
% \and
}

\makeatletter
\g@addto@macro\@maketitle{
    \vspace{-0.5cm}
    \begin{figure}[H]
    \setlength{\linewidth}{\textwidth}
    \setlength{\hsize}{\textwidth}
    \centering
    \includegraphics[width=0.99\linewidth]{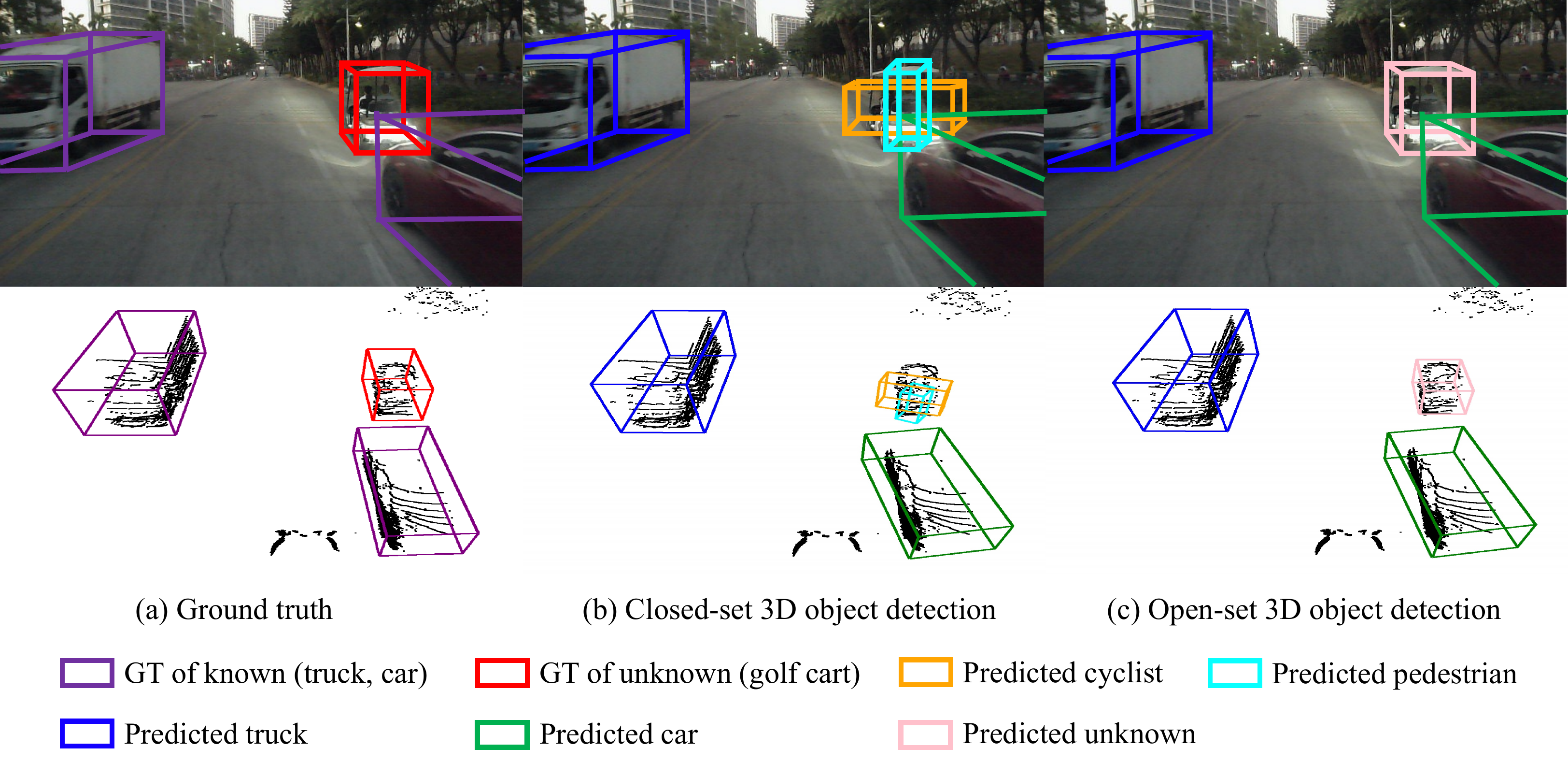}
    \caption{
Illustration of our proposed open-set 3D object detection. GT means ground truth. Classical closed-set object detection can only predict the classes involved during training, so it cannot handle unknown classes, often regarding them as known classes by mistake. The golf cart is not included in the training dataset, and the closed-set detection predicts these points as the pedestrian and cyclist (middle). By contrast, our proposed open-set object detection classifies these points as an unknown class, and gives the accurate bounding box (right).
    }
    \label{fig:system}
    \end{figure}
    % \vspace{-3mm}
}
\makeatother

\maketitle
\thispagestyle{empty}

%%%%%%%%% ABSTRACT
\begin{abstract}
   3D object detection has been wildly studied in recent years, especially for robot perception systems. However, existing 3D object detection is under a closed-set condition, meaning that the network can only output boxes of trained classes. Unfortunately, this closed-set condition is not robust enough for practical use, as it will identify unknown objects as known by mistake. Therefore, in this paper, we propose an open-set 3D object detector, which aims to (1) identify known objects, like the closed-set detection, and (2) identify unknown objects and give their accurate bounding boxes. Specifically, we divide the open-set 3D object detection problem into two steps: (1) finding out the regions containing the unknown objects with high probability and (2) enclosing the points of these regions with proper bounding boxes. The first step is solved by the finding that unknown objects are often classified as known objects with low confidence, and we show that the Euclidean distance sum based on metric learning is a better confidence score than the naive softmax probability to differentiate unknown objects from known objects. On this basis, unsupervised clustering is used to refine the bounding boxes of unknown objects. The proposed method combining metric learning and unsupervised clustering is called the MLUC network. Our experiments show that our MLUC network achieves state-of-the-art performance and can identify both known and unknown objects as expected.
\end{abstract}
%%%%%%%%% BODY TEXT
% \vspace{-0.5cm}
\section{Introduction}

3D object detection plays an important role in many perception systems, such as autonomous driving and robotics. LIDAR is a popular sensor to obtain the 3D point cloud for 3D object detection due to its robustness to the environment. Therefore, various deep learning-based methods~\cite{yan2018second,zhou2018voxelnet,qi2017pointnet,pointnet++,pointrcnn} have been proposed to tackle this point cloud object detection problem. However, classical 3D object detection operates under a closed-set condition, meaning the network only outputs boxes of trained classes. Such a closed-set system will wrongly assign labels of known classes to unknown objects, which could have disastrous consequences in real-world applications~\cite{bozhinoski2019safety}. Therefore, a method for 3D object detection with an open-set condition is needed, as illustrated in Fig.~\ref{fig:system}.

Open-set classification and semantic segmentation tasks are the foundations of the open-set object detection task. Two mainstream approaches to solve the open-set image-level and pixel-level classification problems are uncertainty estimation-based methods~\cite{gal2016dropout,hendrycks2016baseline,DBLP:conf/nips/Lakshminarayanan17} and generative-based methods~\cite{baur2018deep,Lis2019,xia2020synthesize}. However, these methods cannot be adapted to the open-set object detection directly as classification tasks do not have to consider whether an object exists or the location of the object. Recently, open-set 2D object detection has been systematically formulated~\cite{9093355}, and a dropout sampling-based method~\cite{miller2018dropout} and prototypical learning-based method~\cite{joseph2021towards} have been proposed to detect unknown objects. The only previous research that operates under the open-set 3D condition was proposed by Kelvin \etal ~\cite{wong2020identifying}, who use class-agnostic embeddings to cluster unknown objects to solve the open-set 3D instance segmentation task. Inspired by this open-set 2D object detection and 3D instance segmentation, we propose the first open-set 3D object detector, which is able to identify both known and unknown objects in 3D space, using weaker supervision (bounding boxes) instead of point-level annotations.

% The open-set 3D object detector is more suitable than existing close-set 3D object detector to be used in the real world. It is impossible to involve every class in the open world into the training dataset. The perception system of autonomous car and robotics will be more robust if it is able to identify unknown objects. 

Towards the goal of recognizing unknown objects in the LIDAR point cloud, we firstly propose a naive open-set 3D object detector and analyze the difficulty of the open-set 3D object detection task. We find that the naive detector cannot handle the task well, as it mis-classifies known objects as unknown objects, generates lots of false positives of unknown objects, and places inaccurate bounding boxes for unknown objects. To solve these problems, we propose a novel perception system: the Metric learning with Unsupervised Clustering (\textit{MLUC}) network. Specifically, we use metric learning to identify boxes with low confidence scores, and regard these regions as having a high probability of containing unknown objects. On top of this, an unsupervised clustering algorithm is used to generate the precise bounding boxes of unknown objects. In summary, our contributions are the following:
\vspace{-0.2cm}
\begin{itemize}
\setlength{\itemsep}{0pt}
\setlength{\parsep}{0pt}
\setlength{\parskip}{0pt}
    \item We are the first to introduce open-set 3D object detection task, which is more suitable than closed-set for real-world applications such as autonomous driving and robotics.
    \item We analyze the shortcomings of the naive open-set 3D object detector, which straightforwardly extends the closed-set 3D detector to the open-set task.
    \item To solve the problems of the naive open-set 3D object detector, we develop the MLUC network, which combines metric learning and unsupervised clustering to identify both known and unknown 3D objects. We show that our MLUC network achieves state-of-the-art performance compared with other baselines.
\end{itemize}

\section{Related Work}

\subsection{Closed-set 3D Object Detection}

Classical closed-set 3D object detection methods can be divided into two types: grid-based methods and point-based methods. 

Grid-based methods transform irregular point data to regular grids so that the data can be processed by a 2D or 3D convolutional neural network (CNN). MV3D~\cite{chen2017multi} firstly projects the 3D point data to 2D bird's-eye-view grids and then applies a 2D object detection method to generate bounding boxes. Following MV3D, more efficient frameworks with a bird's-eye-view representation are proposed~\cite{yang2018pixor,lang2019pointpillars}. VoxelNet~\cite{zhou2018voxelnet} divides the point clouds into 3D voxels and applies a 3D CNN to extract features, and SECOND~\cite{yan2018second} introduces 3D sparse convolution~\cite{graham20183d} for efficient processing. 

Point-based methods mostly rely on PointNet~\cite{qi2017pointnet} and its variants~\cite{pointnet++,9320379}. PointRCNN~\cite{shi2019pointrcnn} is a typical two-stage point-based 3D object detector, while 3DSSD~\cite{yang20203dssd} introduces F-FPS and is the first one-stage point-based 3D object detector.

These closed-set 3D object detection methods achieve remarkable performance on autonomous driving datasets, such as the KITTI dataset~\cite{geiger2013vision} and Waymo open dataset~\cite{sun2020scalability}. However, they operate under the closed-set condition, and cannot predict unknown objects, which is different to the case in the open world.

\subsection{Open-set 2D Object Detection}
\label{sec:drop}

Dhamija \etal ~\cite{9093355} systematically study the open-set performance of common 2D object detectors~\cite{ren2015faster,lin2017focal,redmon2017yolo9000}, and they find that unknown objects from the open world end up being incorrectly detected as known objects, often with very high confidence. Miller \etal ~\cite{miller2018dropout} use Monte Carlo Dropout~\cite{gal2016dropout} sampling to measure the uncertainty scores of detected boxes, and regard high uncertainty boxes as unknown objects. Recently, Joseph \etal ~\cite{joseph2021towards} adopted prototypical learning with contrastive clustering and an energy-based identifier to detect unknown objects. We draw inspiration from these open-set 2D object detection methods, to propose our MLUC network to address the open-set object detection problem in 3D space.

\subsection{Open-set 3D Instance Segmentation}
\label{sec:osis}

The open-set 3D instance segmentation (OSIS) network, proposed by Kelvin \etal ~\cite{wong2020identifying} is the only previous research to focus on perception under open-set 3D space conditions. The authors use the embedding head to extract the class-agnostic embeddings for each point as well as prototypes for each instance of the known class. During inference, the prototypes collectively filter out points from the known classes whose embeddings are close enough to the prototypes. Then, the remaining points not assigned to any prototypes are clustered into instances of the unknown class based on their embeddings. On top of their OSIS network, we use bounding boxes to enclose the unknown instances, so that this modified OSIS method can be one of the baselines of open-set 3D object detection task. 

%-------------------------------------------------------------------------
\section{Open-set 3D Object Detection}

In this section, we formulate the definition of open-set 3D object detection. We define the class set of the training dataset as $\mathcal{D}^{train} = \left \{ 1,2,...,C\right \}$. In classical closed-set 3D object detection, the class set of the test dataset is the same as that of the training dataset, meaning that $\mathcal{D}^{test} = \mathcal{D}^{train} = \left \{ 1,2,...,C\right \}$. However, in the open-set 3D object detection, we assume that the test dataset contains more categories than the training dataset, which is more close to real applications. Therefore, under the open-set condition, $\mathcal{D}^{test} = \left \{ 1,2,...,C,C+1,C+2,...,C+n\right \} \supseteq \mathcal{D}^{train}$. The label of the training dataset is the bounding boxes set $\mathcal{{B}}^{train}=\left \{ \textbf{b}_{i} | i=1,2,...,m\right \}$, where $m$ refers to the total number of bounding boxes in the training set. Each box can be represented by $\textbf{b}_{i}=[c,x,y,z,w,l,h,\theta]$, where $c\in \mathcal{D}^{train},x,y,z,w,l,h,\theta$ refer to the class, center coordinates, size and rotational angles along $z$ axis of the box. The label of the test set is similar to that of the training set, except that the test set has a larger class set space.

The purpose of the open-set 3D object detection is to train a neural network to not only identify trained $C$ object classes, but also assign the 'unknown' label to those classes not encountered during training that are $\left \{ C+1,C+2,...,C+n\right \}$, as well as use correct bounding boxes to enclose the corresponding points, as shown in Fig.~\ref{fig:system}.

\section{Naive Open-set 3D Object Detector}
\label{sec:naive}

\begin{figure}
\begin{center}
\includegraphics[width=0.99\linewidth]{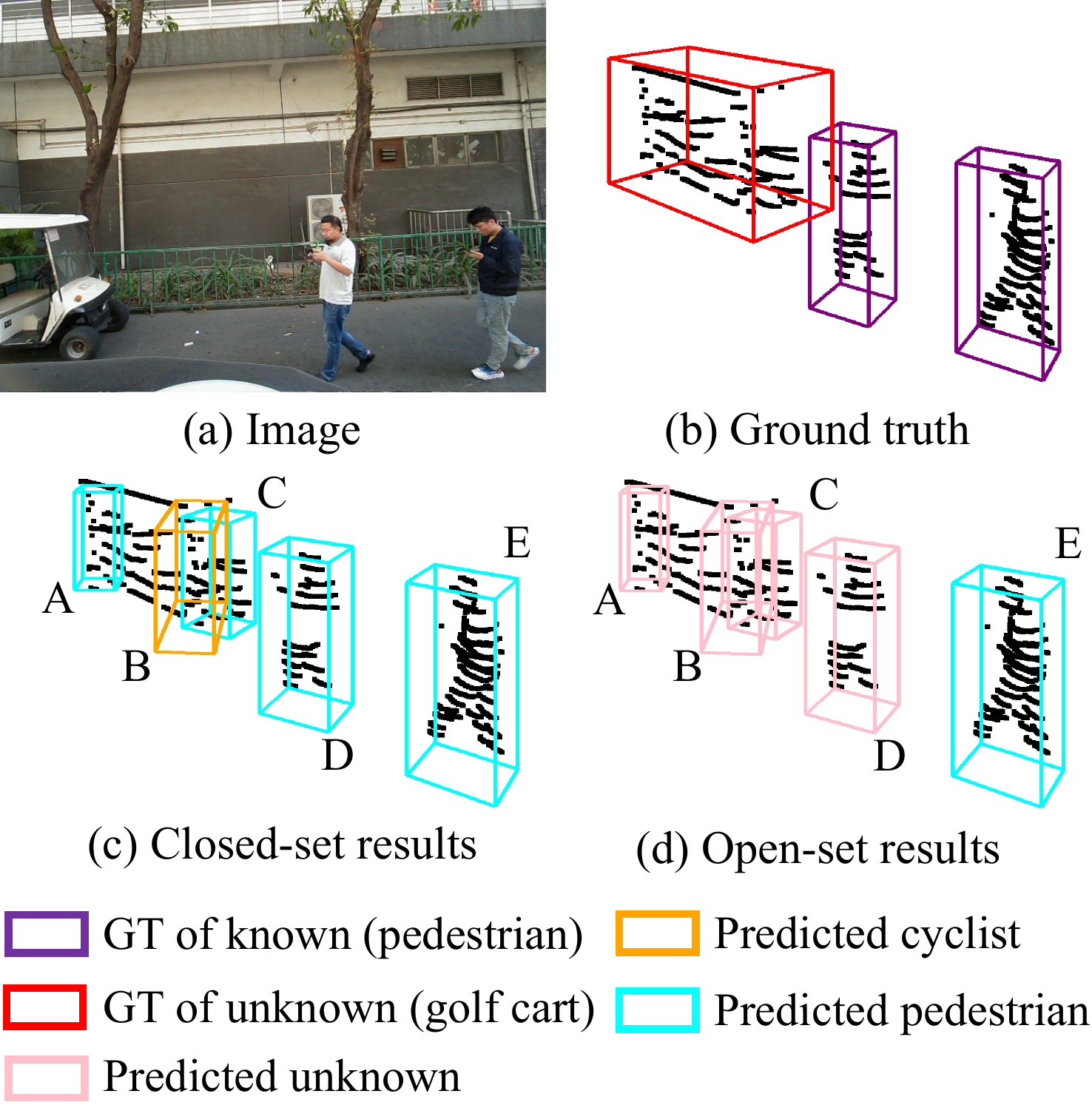}
\end{center}
\vspace{-0.3cm}
   \caption{Prediction example of the naive open-set 3D object detector. Objects A, B, C, and D have low confidence scores, so they are classified as unknown objects by the naive open-set 3D object detector.}
\label{fig:naive}
\end{figure}

Suppose the output of the classical closed-set 3D object detector is $\mathcal{{\hat{B}}}^{close}=\left \{ \mathbf{\hat{b}}_{i} | i=1,2,...,\hat{m}\right \}$, and each $\mathbf{\hat{b}}_{i}=[\hat{c},\hat{s},\hat{x},\hat{y},\hat{z},\hat{w},\hat{l},\hat{h},\hat{\theta}]$, where $\hat{c},\hat{s},\hat{x},\hat{y},\hat{z},\hat{w},\hat{l},\hat{h},\hat{\theta}$ refer to predicted class, confidence score, center coordinates, size and rotational angle of the predicted box. Specifically, the confidence score $\hat{s}$ is determined by:

\vspace{-0.3cm}
\begin{equation}
    \hat{s} = max \left \{ \hat{p}_{i}|i=1,2,...,C\right \},
\end{equation}
where $p_{i}$ refers to the softmax probability of a certain class.

The most natural way to solve the open-set 3D object detection problem is to regard those boxes whose confidence scores are smaller than a threshold ($\hat{s}<\lambda_{naive}$) as unknown objects, which we call the naive open-set 3D object detector. We use this method as one of the baselines. One visualization of the naive open-set 3D object detector is shown in Fig.~\ref{fig:naive}. We can see that although this method is straightforward and simple enough, it has several problems.

\begin{figure*}
\begin{center}
\includegraphics[width=0.99\linewidth]{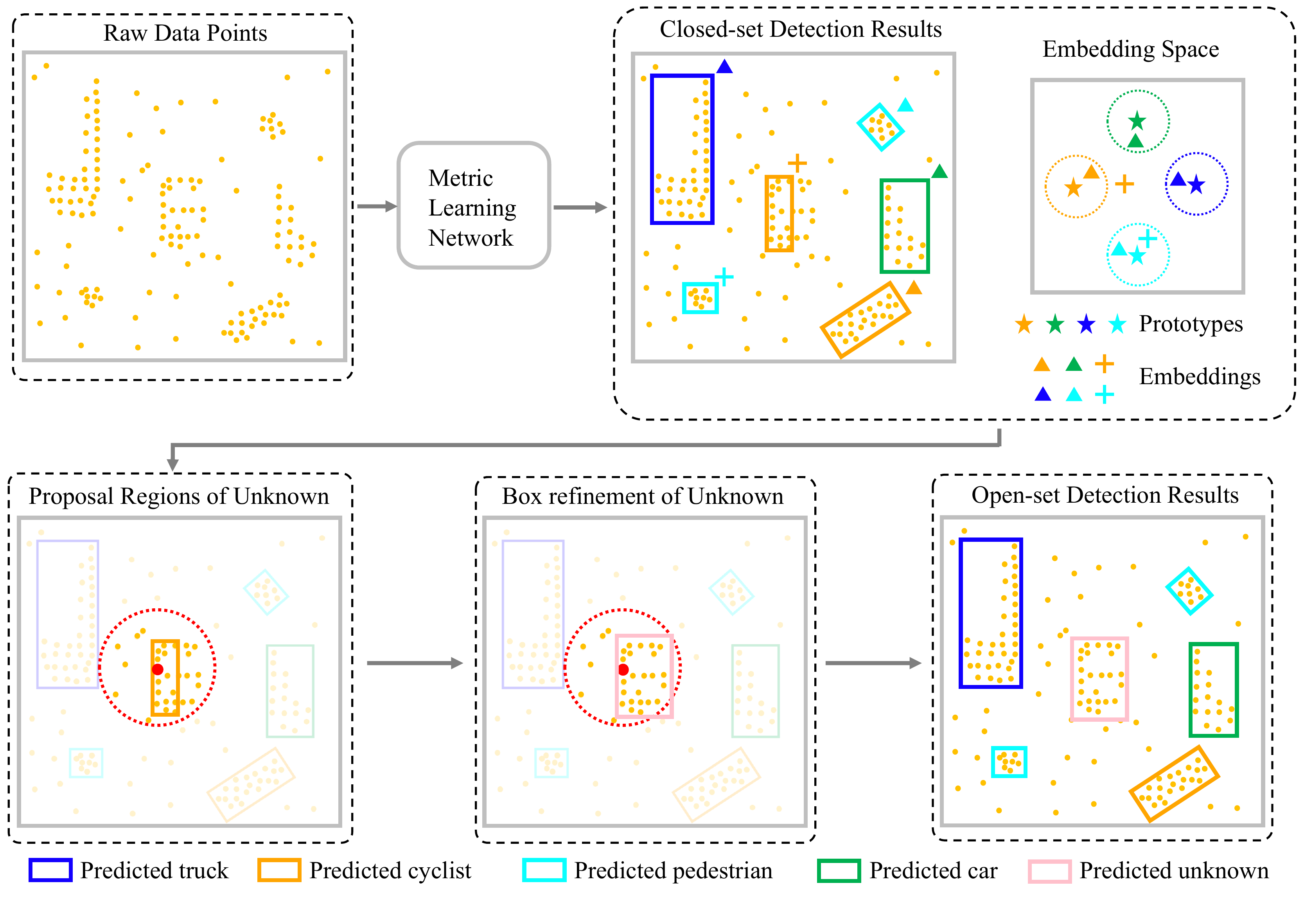}
\end{center}
\vspace{-0.7cm}
   \caption{The pipeline of our MLUC network. A metric learning network is applied for the raw data points to obtain the closed-set detection boxes and corresponding embeddings. Those boxes whose embedding is located at the center of the embedding space are regarded as low-confidence-score boxes. We randomly pick one point from each low-confidence-score box and obtain the proposal regions of unknown objects using picked points as centers. Then unsupervised clustering is used to refine the boxes of unknown objects. In this way, our MLUC network can identify both known and unknown objects to fulfill the open-set 3D object detection task.}
\label{fig:pipeline}
\end{figure*}

\textbf{Mis-classifies known objects as unknown objects.} It is possible to mis-classify some known objects as unknowns~\cite{gal2016dropout,hendrycks2019scaling,DBLP:conf/nips/Lakshminarayanan17}. The reasons include two folds. On one hand, the prediction of unknowns highly depends on closed-set predictions. Since the whole probability space is divided for known objects and there is no remaining space for unknown objects, a foreground object might get mis-classified as unknowns due to a relatively low confidence score or high threshold. On the other hand, only one class, which owns the maximum softmax probability, gets considered in estimating unknown objects. The naive method does not exploit the information of the other classes. In Fig.~\ref{fig:naive}, the known object D is classified as unknown by mistake.

% This is because: (1) The whole probability space is divided for known objects and there is no remaining space for unknown objects. (2) The maximum softmax probability only considers one class explicitly, which is the most likely class. Fig.~\ref{fig:naive} shows that naive method mis-classifies object D as an unknown object.

\textbf{Generates many false positive unknown objects.} In our experiments, we find that the 3D detector tends to classify unknown objects to multiple known objects. For example, the golf cart in Fig.~\ref{fig:naive} is divided into two pedestrians and one cyclist. If we regard all of these objects as unknown objects, more false positive unknown objects will be induced. If we filter these bounding boxes using non-maximum-suppression (NMS) with negative softmax probability score, it will make the bounding boxes inaccurate, which will be discussed in the next paragraph.

\textbf{Generates inaccurate bounding boxes for unknown objects.} The size of the predicted boxes from the naive method is very close to the pre-defined anchor size. Therefore, if we only change the labels of some boxes to 'unknown' but keep the size unchanged, the corresponding boxes cannot enclose the points of unknown objects very well. For example, in Fig.~\ref{fig:naive}, the boxes of the pedestrians and cyclists do not match the size of the golf cart. This is why open-set object detection is more difficult than open-set classification.

\section{MLUC Network}

To solve the problems of the naive open-set 3D object detector, we propose the MLUC network. We use the Euclidean distance sum in metric space to measure the uncertainty, and we show that it is a better confidence score compared with the naive maximum softmax probability. Therefore, metric learning helps us find more reliable regions containing unknown objects. On this basis, unsupervised clustering is applied on each proposal region of unknown objects, so that the bounding boxes of unknowns get well refined. Finally, bounding boxes generated for unknown objects will be processed by NMS to filter out overlapping results. In this way, we can obtain more accurate bounding boxes for unknown objects and suppress false positives. The pipeline of the MLUC network is shown in Fig.~\ref{fig:pipeline}.

\begin{figure*}
\begin{center}
\includegraphics[width=0.99\linewidth]{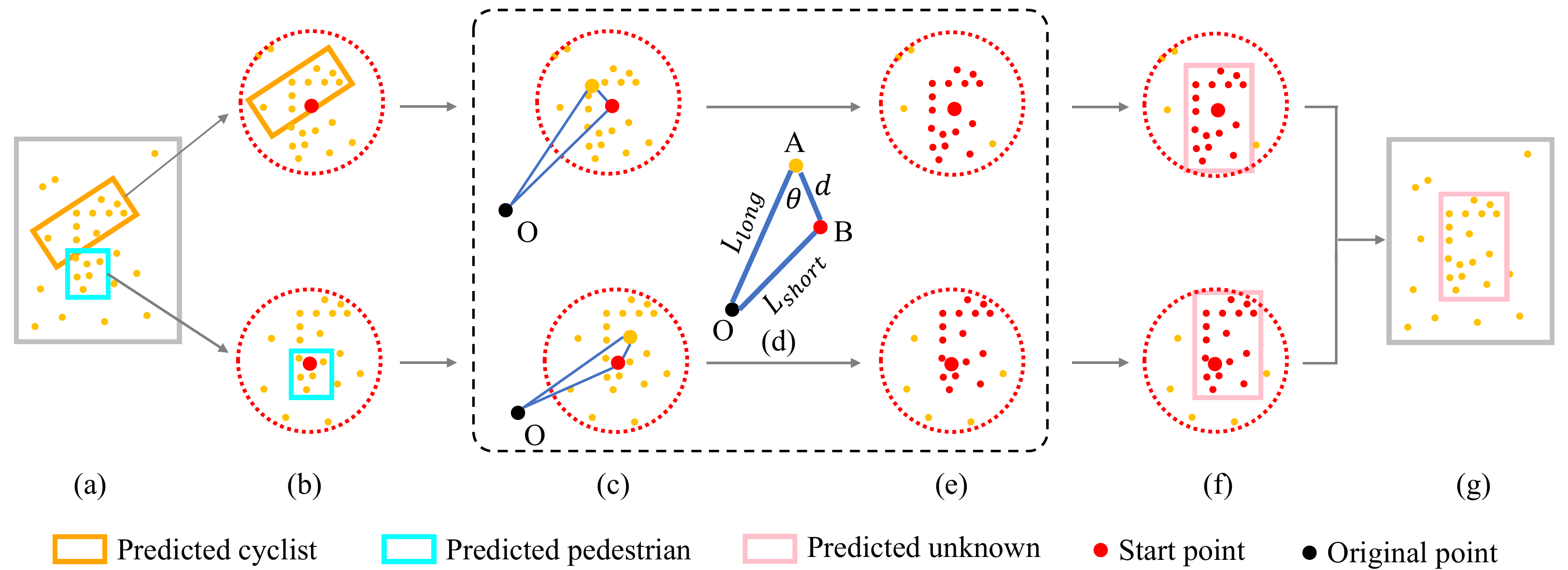}
\end{center}
\vspace{-0.3cm}
   \caption{Illustration of our unsupervised clustering algorithm. (a): Closed-set detection results. (b): Proposal regions of unknown objects. (c): Depth clustering from the start point. (d) Depth clustering mechanism. (e) Depth clustering results. (f) Using a new bounding box to enclose the clustering results. (g) Predicted unknown objects.}
\label{fig:uns}
\end{figure*}

\subsection{Deep Metric Learning}

The classification branch of a typical object detection network is composed of two parts: a feature extractor to obtain the high-dimensional features and a classifier to generate the decision hyper-plane. However, this feature extractor and classifier structure is not suitable for unknown objects detection, as the network assumes that all feature space is assigned for known objects, and there is no space left for unknown objects~\cite{yang2018robust}. Therefore, we replace the classifier by the Euclidean distance representation with all prototypes $\mathcal{M}_{in}=\left \{\mathbf{m}_t\in \mathbb{R}^{1 \times C}|t\in \left \{ 1,2,...,C \right \} \right \}$, where $\mathbf{m}_t$ refers to the prototype of class $\mathcal{D}_{t}^{train}$. The feature extractor $f(\mathbf{X})$ is designed to learn the embedding vector in the metric space of each input box $\mathbf{X}$. In this way, the probability of box $\mathbf{X}$ classified as the class $\mathcal{D}_{t}^{train}$ is formulated as:
\begin{equation}
    p_t(\mathbf{X})=\frac{exp(-{\left \| f(\mathbf{X})-\mathbf{m}_t \right \|}^2)}{\sum_{t'=1}^{C}exp(-{\left \| f(\mathbf{X})-\mathbf{m}_{t'} \right \|}^2)}. \label{con:probability}
\end{equation}
Then we can apply this Euclidean distance-based probability to define the loss function:
\begin{equation}
\small
    \mathcal{L} =  \sum -log(\frac{exp(-{\left \| f(\mathbf{X})-\mathbf{m}_{\mathbf{Y}} \right \|}^2)}{\sum_{k=1}^{C}exp(-{\left \| f(\mathbf{X})-\mathbf{m}_{k} \right \|}^2)}),
\end{equation}
where $\mathbf{Y}$ is the ground truth class of the input box $\mathbf{X}$. This loss function has two effects on the learned embedding vectors: (1) The embedding vector will be attracted by the prototype of the same class, which is affected by the numerator of the loss function. (2) The embedding vector will be repelled by the prototypes of other classes, which is affected by the denominator of the loss function. In this way, the embeddings of known objects will be close to the corresponding prototypes of the same class, while the embeddings of unknown objects will be distributed in the center of the embedding space as they are repelled by all known prototypes, as illustrated in Fig.~\ref{fig:pipeline}.

Based on this metric learning framework, we propose to use the Euclidean distance sum (EDS) to measure the uncertainty. The EDS is defined as:
\begin{equation}
    EDS=\sum_{t=1}^{C}{\left \| f(\mathbf{X})-\mathbf{m}_t \right \|}^2.
\end{equation}

Unknown objects are supposed to have smaller a EDS as they are in the center of the embedding space. Compared to the maximum softmax probability, this EDS considers all classes explicitly. The boxes whose EDS score is smaller than a threshold $(EDS<\lambda_{EDS})$ are considered to contain unknown objects. As EDS is class-independent, all prototypes of known classes are designed to be evenly distributed in the embedding space and stable during training. We define the prototype in a one-hot vector form: only the $t^{th}$ element of $\mathbf{m}_t$ is $C$, while others remain zero, where $t\in \left \{1,2,...,C \right \}$~\cite{miller2020class}.

\subsection{Unsupervised Clustering}

After we obtain the low-confidence-score bounding boxes, we regard them as the proposal regions of unknown objects and apply unsupervised clustering to refine them. Fig.~\ref{fig:uns} shows the process of our unsupervised clustering algorithm.

The first step is to find the proposal regions of unknown objects based on the obtained EDS of each predicted box. Those boxes whose EDS is smaller than a threshold $\lambda_{EDS}$ are considered to have high a probability of containing unknown objects. Then we randomly pick one point from each low EDS box, and set up the proposal regions of unknown objects with the cylinders whose centers are picked points and radius is $r$, as shown in Fig.~\ref{fig:uns} (a) and (b).

The second step is to cluster points of unknown objects in the proposal regions. We execute the depth clustering~\cite{7759050} algorithm to find the points of unknown objects. The principle of depth clustering is illustrated in Fig.~\ref{fig:uns} (d). O is the original point, which is also the location of the LIDAR. A and B are two points for which it is to be decided whether they belong to the same object or not. The point which is far away from O is A and the other is B. If $\theta$ is smaller than a threshold $\lambda_{\theta}$, A and B are considered to belong to one object. We start iterations from the center of the proposal regions, and the points which are seen to belong to the same object with the start point will be the start point of the next iteration, until all points in the proposal region are decided. The red points in Fig.~\ref{fig:uns} (e) are considered to be the points of one unknown object.

The third step is to use a tight bounding box to enclose the points of unknown objects, and then use NMS to post-process all obtained bounding boxes of unknown objects, with larger bounding boxes having higher priority, as shown in Fig.~\ref{fig:uns} (f) and (g).

We summarize our unsupervised clustering method in Algorithm~\ref{algorithm}. $\mathcal{{\hat{B}}}^{close}=\left \{ \mathbf{\hat{b}}_{i} | i=1,2,...,\hat{m}\right \}$ represents the output closed-set bounding boxes of the metric learning, and $\mathbf{\hat{b}}_{i}[\hat{s}]$ is the corresponding EDS value.

\SetAlCapFnt{\small}
\SetAlFnt{\small}
\SetAlTitleFnt{\small}
\begin{algorithm}
% \SetAlFnt{\tiny\sf}
\caption{Unsupervised Clustering Method}\label{algorithm}
\KwInput{$\mathcal{\hat{B}}^{close}=\left \{ \mathbf{\hat{b}}_{i}| i=1,2,...,\hat{m}\right \}$}
\KwOutput{\text{Bounding boxes of unknown objects $\mathcal{\hat{B}}^{unknown}$}}
$P=[ \ ]$\;
\For{$\mathbf{\hat{b}}_{i}$ \KwIn $\mathcal{\hat{B}}^{close}$}
{\uIf{$\mathbf{\hat{b}}_{i}[\hat{s}] < \lambda_{EDS}$}
{randomly pick one point $p$ inside $\mathbf{\hat{b}}_{i}$\;
$P$.append($p$)\;
}}
\For{$p$ \KwIn $P$}
{$Q=[p]$;
$\bar{Q}=R=[ \ ]$\;
\While{$Q$ is not empty}
{$t$=$Q$.top();
$Q$.pop();
$\bar{Q}$.append($t$);
$R$.append($t$)\;
$N_t=$ the neighbor point set of $t$\;
\For{$s$ \KwIn $N_t$}{\uIf{$s$ \KwIn $\bar{Q}$ \KwOr $sp > r$}{continue}
$L_{long}=max(Os,Ot)$\;$L_{short}=min(Os,Ot)$\;$d=st$\;
$\theta=arccos(\frac{L_{long}^2+d^2-L_{short}^2}{2dL_{long}})$\;
\eIf{$\theta<\lambda_{\theta}$}{$Q$.append($s$)}{$\bar{Q}$.append($s$)}}
}
$\mathbf{\hat{b}}^{unknown}=$ the bounding box enclosing $R$ tightly\;
$\mathcal{\hat{B}}^{unknown}$.append($\mathbf{\hat{b}}^{unknown}$)\;
}
$\mathcal{\hat{B}}^{unknown}$=NMS($\mathcal{\hat{B}}^{unknown}$);
\end{algorithm}

\begin{figure*}
\begin{center}
\includegraphics[width=0.99\linewidth]{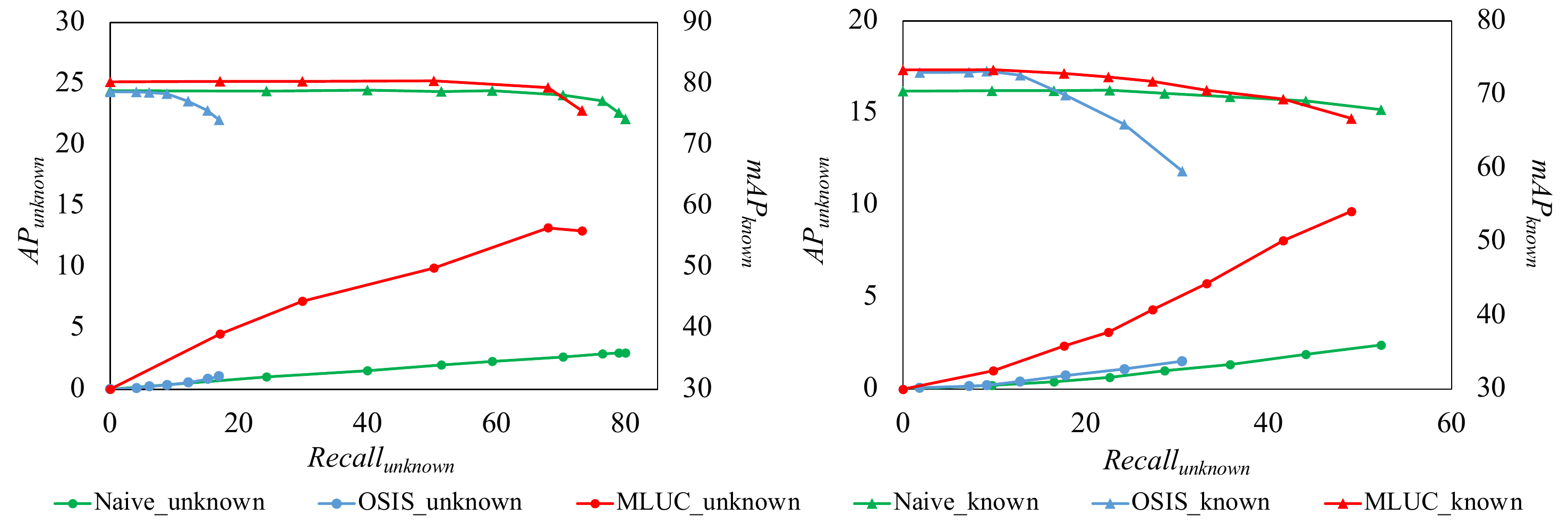}
\end{center}
\vspace{-0.3cm}
   \caption{The performance of Naive, OSIS and MLUC methods on UDI (left) and KITTI (right) dataset. For each confidence score threshold, we can obtain a data point $(Recall_{unknown},AP_{unknown},mAP_{known})$. The two graphs contain the results of $AP_{unknown}$ and $mAP_{known}$ with regard to $Recall_{unknown}$.}
\label{fig:res}
\end{figure*}

\section{Experiments}

% In this section, we first introduce the experimental setup of the open-set 3D object detection and then discuss the results to show the effectiveness of our MLUC network.

\subsection{Experimental Setup}

\noindent \textbf{Datasets:} We evaluate the performance of our MLUC network on two datasets.

\textit{UDI dataset} is a self-driving dataset with LIDAR point clouds collected from an industrial park. Six classes including car, pedestrian, cyclist, truck, golf cart, and forklift, are annotated in the UDI dataset. In our experiments, four classes, car, pedestrian, cyclist, and truck, are treated as known objects, while the other two classes, golf cart, and forklift, are regarded as unknown objects and not involved during training. There are a total 200k known objects in the training set and 12k known objects as well as 600 unknown objects in the test set.

\textit{KITTI dataset}~\cite{geiger2013vision} is one of the most popular open-source datasets of 3D object detection for autonomous driving. Three common classes, car, pedestrian, and cyclist, are used as known objects, while the van and truck are used as the unknown classes and not included during training. There are 8690 training objects and 6100 test objects in our experiments, with the test objects composed of 4845 known objects and 1255 unknown objects.
\vspace{0.3cm}

\noindent \textbf{Evaluation metrics:} For known objects, we use the 3D \textit{mean average precision ($mAP_{known}$)} to evaluate the performance, while for unknown objects, we report the \textit{$recall_{unknown}$} and 3D \textit{average precision ($AP_{unknown}$)}. Then, the metric $mAP_{harm}$ is used to comprehensively evaluate the overall performance~\cite{Xian_2019_CVPR}:
\begin{equation}
    mAP_{harm}=\frac{2*mAP_{known}*AP_{unknown}}{mAP_{known}+AP_{unknown}},
\end{equation}
% \vspace{0.1cm}

\noindent \textbf{Baselines:} We adopt the naive open-set 3D object detector, MC-Dropout, and modified OSIS network, which have been discussed in Section~\ref{sec:naive}, Section~\ref{sec:drop}, and Section~\ref{sec:osis} respectively, as our baselines.
\vspace{0.3cm}

\noindent \textbf{Implementation details:} We adopt \textit{SECOND}~\cite{yan2018second} as our 3D object detector for both the UDI and KITTI dataset.

For the UDI dataset, the detection range of the point clouds is $[-51.2,51.2] \ m$ for both the $X$ and $Y$ axis, and $[-5,3] \ m$ for the $Z$ axis. We use the ADAM~\cite{kingma2015adam} optimizer with learning rate 0.003 and momentum 0.9. The network is trained on one NVIDIA 2080Ti for 20 epochs with a batch size 4. The IoU threshold during evaluation is 0.5 for car, pedestrian, and cyclist, 0.7 for truck, and 0.1 for the unknown objects, i.e., golf cart and forklift. The evaluation difficulty is easy.

For the KITTI dataset, the detection range is $[0,70.4] \ m$ for the $X$ axis, $[-40,40] \ m$ for the $Y$ axis, and $[-3,1] \ m$ for the $Z$ axis. The learning process setting is the same as the UDI dataset except that we train the model for 80 epochs on the KITTI dataset. The IoU threshold during evaluation is 0.7 for car and truck, 0.5 for pedestrian and cyclist, and 0.1 for the unknown objects, i.e., van and truck. The evaluation difficulty is moderate.

The $\lambda_{\theta}$ in the unsupervised clustering is $65^{\circ}$, and $r$ is $4 \ m$ and $5 \ m$ for the UDI and KITTI dataset respectively.

\subsection{Results}

The specific open-set 3D object detection results are related to the confidence thresholds ($\lambda_{naive}$ for Naive method and OSIS method, and $\lambda_{EDS}$ for MLUC method). Therefore, we plot the results for the UDI and KITTI dataset of $AP_{unknown}$ and $mAP_{known}$ with regard to $Recall_{unknown}$ in Fig.~\ref{fig:res}, where each data point is the result of one specific threshold. Fig.~\ref{fig:res} shows the $mAP_{known}$ reduces with the growth of $Recall_{unknown}$, because more known objects are regarded as unknown objects with the growth of the confidence score threshold. So we only record the points whose $mAP_{known}$ is not reduced by $10\%$ to ensure the high performance of the known classes. Then we pick the result with the best $mAP_{harm}$ to represent the optimal performance of each method, and these are recorded in Table~\ref{tab:1}. 

\begin{figure*}
\begin{center}
\includegraphics[width=0.99\linewidth]{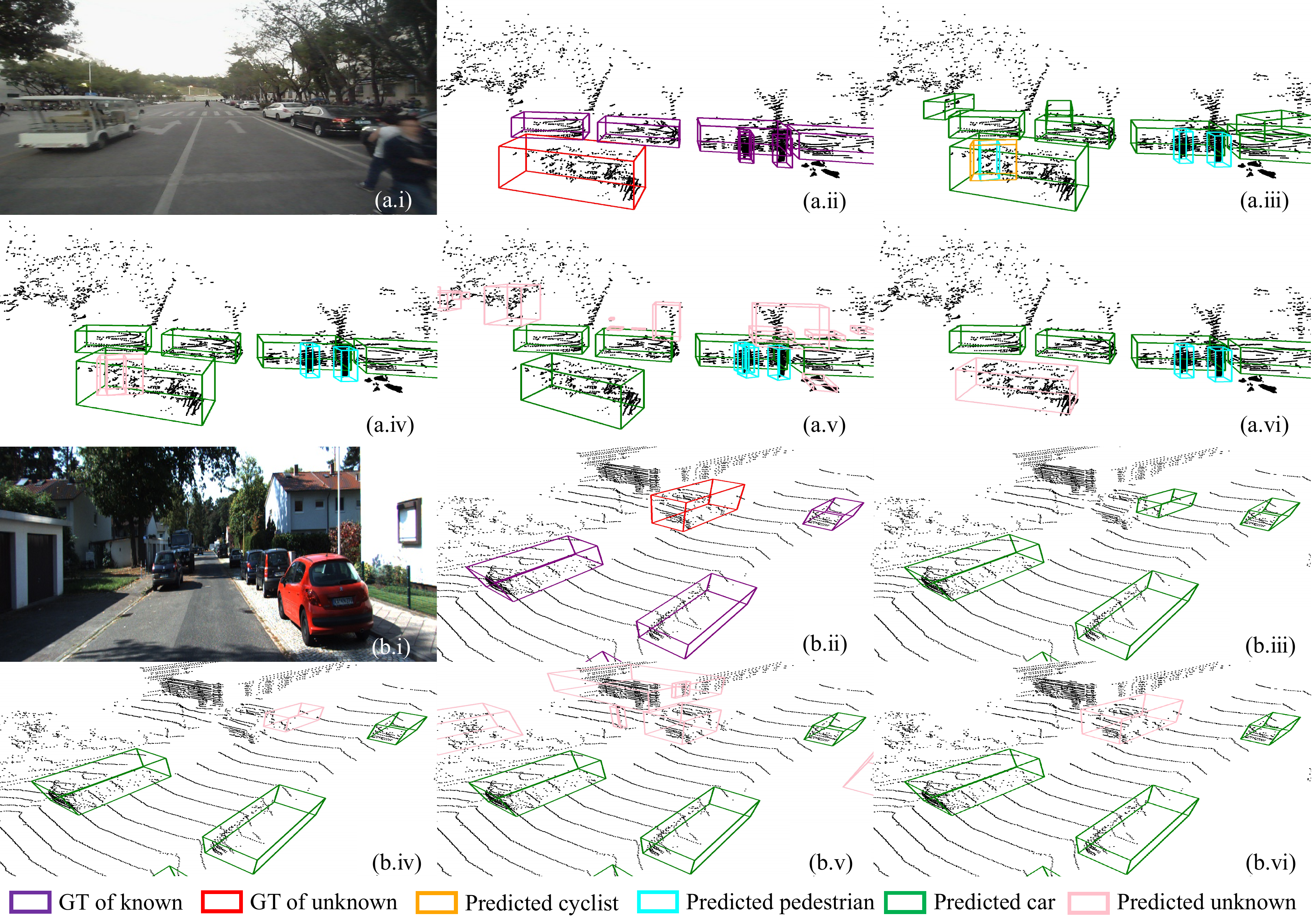}
\end{center}
\vspace{-0.3cm}
   \caption{Visualization of some qualitative results. (a) UDI dataset; (b) KITTI dataset; (\romannumeral1) Image; (\romannumeral2) Ground truth; (\romannumeral3) Closed-set results; (\romannumeral4) Naive method; (\romannumeral5) OSIS method; (\romannumeral6) MLUC method. Golf cart and truck are unknown objects for UDI and KITTI dataset respectively. (a.\romannumeral3) and (b.\romannumeral3) show that unknown objects are classified as known objects by mistake. The Naive method cannot provide accurate bounding boxes for unknown objects according to (a.\romannumeral4) and (b.\romannumeral4). The OSIS method regards other stuff things including trees and walls as unknown objects, as shown in (a.\romannumeral5) and (b.\romannumeral5). Our MLUC method can provide accurate locations and boxes of unknown objects, as indicated in (a.\romannumeral6) and (b.\romannumeral6).}
\label{fig:exp}
\end{figure*}

Fig.~\ref{fig:res} and Table~\ref{tab:1} show that our MLUC method has the best $AP_{unknown}$ and $mAP_{harm}$ compared with the Naive method and OSIS method. From Fig.~\ref{fig:res}, we also find that the Naive method and MLUC method have a larger $Recall_{unknown}$ than the OSIS method when $mAP_{known}$ decreases within $10\%$. This is because an unknown object is often classified as several overlapping known objects, as shown in Fig.~\ref{fig:system} and~\ref{fig:naive}, and in the OSIS method, points will only be considered to be part of unknown objects when all boxes that include them have higher confidence scores than the threshold. In contrast, the Naive method and MLUC method only require the lowest score of these overlapping boxes to be higher than the threshold. Two visualization results are shown in Fig.~\ref{fig:exp}.

To validate the effectiveness of the metric learning, EDS score, and unsupervised clustering in the MLUC method, we conduct ablation experiments and show the results in Table~\ref{tab:2}. It shows all these three components make a contribution to the better open-set detection performance.

% \begin{table}[!ht]
% \begin{center}
% \begin{tabular}{ccccc}
% \toprule [1pt]
% \multicolumn{5}{c}{UDI dataset} \\ \hline
% ML & UC & $mAP_{known}$ & $AP_{unknown}$ & $mAP_{harm}$   \\ \hline
%   \small{\xmark} & \small{\xmark}    & 75.3 & 3.0  & 5.7  \\
% \checkmark   & \small{\xmark}    & 78.7 & 8.3  & 15.1 \\
% \checkmark   & \checkmark   & \textbf{79.4} & \textbf{13.2} & \textbf{22.6} \\ \midrule [1pt]
% \multicolumn{5}{c}{KITTI dataset} \\ \hline
% ML & UC & $mAP_{known}$ & $AP_{unknown}$ & $mAP_{harm}$   \\ \hline
% \small{\xmark}   &\small{\xmark}& 63.8 & 3.9  & 7.3  \\
% \checkmark   & \small{\xmark}    & 66.1 & 5.8  & 10.6 \\
% \checkmark   & \checkmark   & \textbf{66.8} & \textbf{9.7} & \textbf{16.9} \\
% \bottomrule [1pt]
% \end{tabular}
% \end{center}
% \caption{Ablation experiment results of MLUC method. ML and UC refer to metric learning and unsupervised clustering respectively.}
% \label{tab:2}
% \end{table}

\begin{table}[!ht]
\small
\renewcommand\arraystretch{1}
\renewcommand\tabcolsep{4pt}
\begin{center}
\begin{tabular}{cccccc}
\toprule [1pt]
\multicolumn{6}{c}{UDI dataset} \\ \hline
ML & EDS & UC & ${mAP_{known}}$ & $AP_{unknown}$ & $mAP_{harm}$   \\ \hline
  \small{\xmark} & \small{\xmark} & \small{\xmark}    & 75.3 & 3.0  & 5.7  \\
\checkmark   & \small{\xmark} & \small{\xmark}    & 77.6 & 5.2  & 9.8 \\
\checkmark   &\checkmark & \small{\xmark}    & 78.7 & 8.3  & 15.1 \\
\checkmark   & \checkmark   & \checkmark & \textbf{79.4} & \textbf{13.2} & \textbf{22.6} \\ \midrule [1pt]
\multicolumn{6}{c}{KITTI dataset} \\ \hline
ML & EDS & UC & $mAP_{known}$ & $AP_{unknown}$ & $mAP_{harm}$   \\ \hline
\small{\xmark}   &\small{\xmark} & \small{\xmark} & 63.8 & 3.9  & 7.3  \\
\checkmark   & \small{\xmark} & \small{\xmark}    & 65.9 & 4.7  & 8.8 \\
\checkmark   & \checkmark & \small{\xmark}    & 66.1 & 5.8  & 10.6 \\
\checkmark   & \checkmark   &\checkmark & \textbf{66.8} & \textbf{9.7} & \textbf{16.9} \\
\bottomrule [1pt]
\end{tabular}
\vspace{-0.3cm}
\end{center}
\caption{Ablation experiment results of MLUC method. ML, EDS, and UC refer to metric learning, EDS score, and unsupervised clustering respectively.}
\label{tab:2}
\end{table}

\begin{table*}[t]
\begin{center}
\begin{tabular}{l|ccc|ccc}
\toprule [1pt]
Dataset    & \multicolumn{3}{c|}{UDI}                         & \multicolumn{3}{c}{KITTI}                        \\ \midrule
Methods     & $mAP_{known}$ & $AP_{unknown}$ & $mAP_{harm}$ & $mAP_{known}$ & $AP_{unknown}$ & $mAP_{harm}$ \\ \midrule
Closed-set  & 78.9           & 0               & 0             &  70.5              & 0                 & 0               \\
Supervised &  78.4              &  61.7               &     69.1          & 76.6               &  80.6               &  78.5             \\ \midrule
Naive      & 75.3           & 3.0             & 5.7           & 63.8               &  3.9               &  7.3             \\
MC-Dropout  & 75.9 & 0.1 & 0.2 & 64.1 & 2.6 & 5.0 \\
OSIS       & 74.1                & 1.1                & 2.1              & 65.9               & 1.1                &    2.2           \\ 
MLUC       & \textbf{79.4}           & \textbf{13.2}            & \textbf{22.6}          & \textbf{66.8}               & \textbf{9.7}                & \textbf{16.9}             
\\ 
\bottomrule [1pt]
\end{tabular}
% \vspace{0.5cm}
\end{center}
\caption{Optimal performance of open-set 3D object detection. Supervised method means we include the unknown classes in the training set and retrain the model, so it is the upper bound of the open-set detection performance. We show that our MLUC method achieves the best performance among all baselines.}
\label{tab:1}
\end{table*}
\vspace{-0.5cm}

\section{Conclusion}

In this paper, we propose open-set 3D object detection to identify both known and unknown objects in 3D LIDAR point clouds. We show that our metric learning with unsupervised clustering (MLUC) method achieves the best performance compared with the Naive method and OSIS method on the UDI and KITTI datasets. The performance gap between our MLUC method and supervised upper bound indicates that this open-set 3D object detection problem can be further studied. We hope our work can draw more researchers to contribute to this practically valuable research direction.

{\small
\bibliographystyle{unsrt}
\bibliography{egbib}
}

\end{document}